\documentclass{article}
\usepackage{ijcai20}

\usepackage{times}
\usepackage{soul}
\usepackage{url}
\usepackage {color}
\usepackage[utf8]{inputenc}
\usepackage[small]{caption}
\usepackage{graphicx}
\usepackage{amsmath}
\usepackage{amsthm}
\usepackage{booktabs}
\usepackage{adjustbox}
\usepackage{multirow}
\usepackage{subcaption}
\usepackage{arydshln}
\newcommand{\newcite}[1]{\citeauthor{#1}~\shortcite{#1}}
\newcommand{\citep}[1]{\citeauthor{#1},~\citeyear{#1}}
\usepackage[ruled,linesnumbered]{algorithm2e}

\title{CoSDA-ML: Multi-Lingual Code-Switching Data Augmentation  for Zero-Shot Cross-Lingual NLP}

\author{
	Libo Qin$^{1}$
	\and
	Minheng Ni$^{1}$\and
	Yue Zhang$^{2,3}$\and Wanxiang Che$^{1}$
	\affiliations
	$^1$Research Center for Social Computing and Information Retrieval \\Harbin Institute of Technology, China\\
	$^2$School of Engineering, Westlake University, China\\
	$^3$Institute of Advanced Technology, Westlake Institute for
	Advanced Study
	\emails
\{lbqin, mhni, car\}@ir.hit.edu.cn,
	yue.zhang@wias.org.cn
}

\begin{document}

\maketitle

\begin{abstract}
	Multi-lingual contextualized embeddings, such as multilingual-BERT (mBERT), have
	shown success in a variety of zero-shot cross-lingual tasks.
	However, these models are limited by having inconsistent contextualized representations of subwords across different languages.
	Existing work addresses this issue by bilingual projection and fine-tuning technique.
	We propose a data augmentation framework to generate multi-lingual code-switching data to fine-tune mBERT, which encourages model to align representations from source and multiple target languages once by mixing their context information.
	Compared with the existing work, our method does not rely on bilingual sentences for training, and requires only one training process for multiple target languages.
 	Experimental results on five tasks with 19 languages show that our method leads to significantly improved performances for all the tasks compared with mBERT.
% 	In addition, comprehensive analysis empirically shows the feasibility of our method.
\end{abstract}

\section{Introduction}
Neural network models for NLP rely on the availability of labeled data for effective training \cite{yin2019dialog}. For languages such as English and Chinese, there exist manually labeled datasets on a variety of tasks, trained over which neural models for NLP can rival human performance. However, for most of the languages, manually labeled data can be scarce. As a result, cross-lingual transfer learning stands as a useful research task \cite{ruder2017survey}. The main idea is to make use of knowledge learned from a resource-rich language to enhance model performance on a low-resource language. In particular, zero-shot cross-lingual learning has attracted much research attention \cite{wang-etal-2019-cross}, which requires no labeled data for a target language. In this paper, we consider this transfer setting.

%While traditional methods have considered ... ... and ..., 
Recent state-of-the-art results have been achieved by methods based on cross-lingual contextualized embeddings \cite{conneau2019cross,huang-etal-2019-unicoder,liu2019attentioninformed,devlin-etal-2019-bert}. In particular, a common set of subwords are extracted across different languages, which are taken as the basis for training contextualized embeddings. For such training, raw sentences from multiple languages are merged into a single training set, so that shared subword embeddings and other parameters are tuned across different languages. A representative model is mBERT \cite{devlin-etal-2019-bert}, which is a multi-lingually trained version of BERT.
\begin{figure}[t!]
	\centering
	\includegraphics[width=0.95\columnwidth]{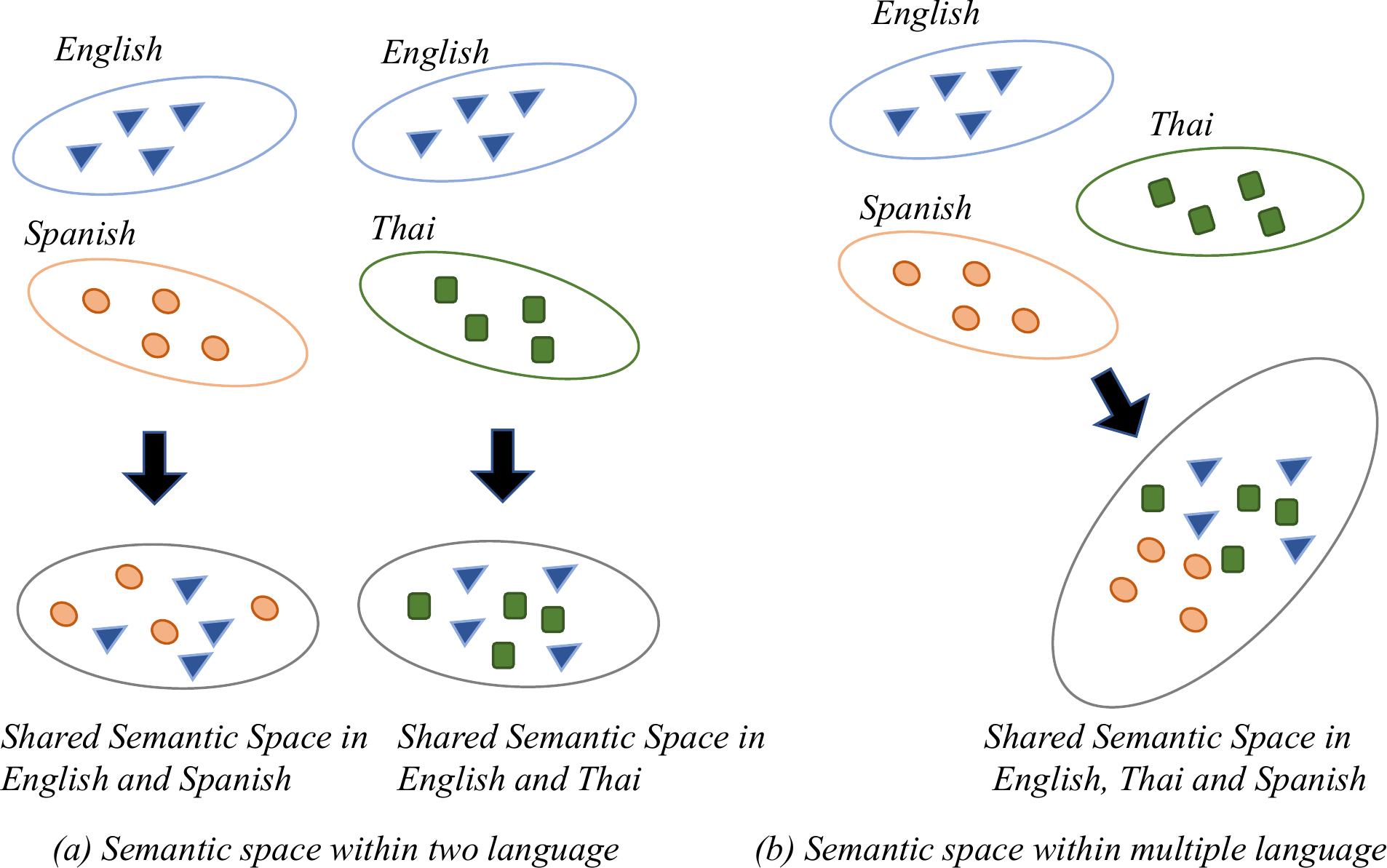}
	\caption{Prior work (a) vs. our method (b).}\label{fig:align}
\end{figure}

While the method above gives strong results for zero-shot cross-lingual adaptation through shared subwords and parameters, it has a salient limitation.  The context for training cross-lingual embeddings is still mono-lingual, which can lead to inconsistent contextualized representations of subwords across different languages. To address this issue, several recent methods try to bridge the inconsistency of contextualized embeddings across languages. As shown in Figure~\ref{fig:align}(a), two main lines of methods are considered. One learns a mapping function from a source contextualized subword embedding to its target counterpart by using word alignment information \cite{wang-etal-2019-cross}, and the other uses code mixing to construct training sentences that consist of both source and target phrases in order to fine-tune mBERT \cite{liu2019attentioninformed}. 
Unfortunately, 
both lines of work only consider a pair of source and target languages at a time, therefore resulting in a separate model for each target language.

\begin{figure}[t!]
	\centering
	\includegraphics[width=0.9\columnwidth]{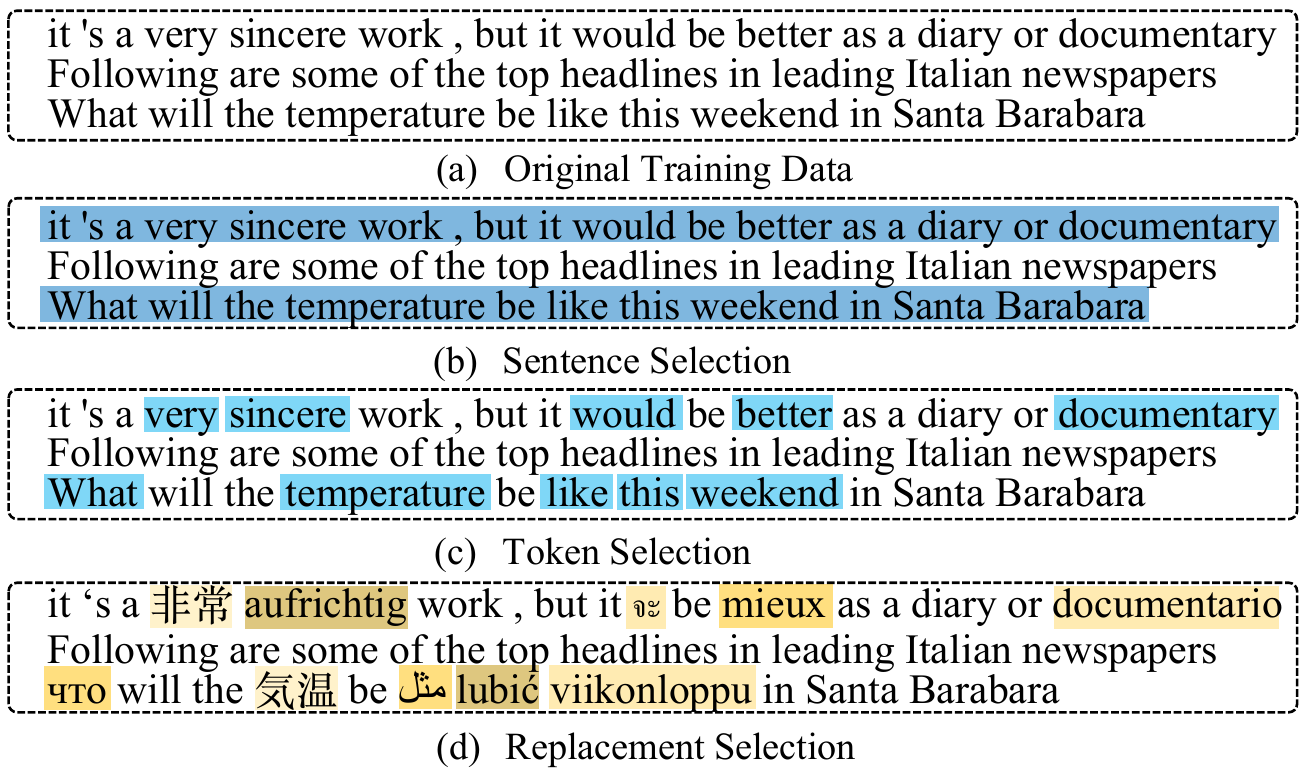}
	\caption{Augmentation process. The source language sentences (a), the sentence selection step (b), the token selection step (c) and the replacement selection step (d) (different shades yellow colors in (d) represent different languages translation). }\label{fig:example}
\end{figure}

We consider enhancing mBERT without creating multiple additional models, by constructing code-switched data in multiple languages dynamically for better fine-tuning. To this end, a set of English raw sentences and the bilingual dictionaries of MUSE \cite{lample2018word} are used as the basis. \footnote{For some languages that can not be found in MUSE, we extract their dictionaries from Panlex \cite{kamholz-etal-2014-panlex}.}  As shown in Figure~\ref{fig:example}, three data augmentation steps are taken. First, a set of sentences is randomly selected for code mixing. Second, a set of words is randomly chosen in each sentence for being replaced with the translation words in a different language. Third, for each word to translate, a target language is randomly selected. The above procedure is dynamically executed on a batch level, for fine-tuning mBERT.
The intuition is to help the model automatically and implicitly align the replaced word vectors in the source and all target languages by mixing their context information.

Compared with existing methods, our method has the following advantages. First, the resulting model is as simple to use as mBERT, without the need to know the test language before hand.
In addition, one training process is used for all different target languages.
 Second, unlike most existing methods, our method does not rely on parallel sentences, which is especially practical for low-resource languages.
%  In addition, one training process is used for all different target languages. 
%This advantage is demonstrated by Figure~\ref{fig:align} (b). 
Third, the method is dynamic in the sense that a different set of code-switched sentences is constructed in each batch during training, therefore increasing the distribution of data instances \cite{liu2019roberta}. Finally, contextualized embeddings for all the languages are aligned into the same space while prior work can only align representation in source and one target language for each language.
This advantage is demonstrated in Figure~\ref{fig:align} (b). 

We conduct experiments on five zero-shot cross-lingual tasks: natural language inference, sentiment classification, document classification, natural language understanding and dialogue state tracking. Results show that our method leads to significantly improved performances for all the tasks compared with mBERT.
% In addition, our method also outperforms the existing enhancement methods over multi-lingual contextualized representation methods mentioned earlier. 
%Besides, we find that our method is particularly helpful for small datasets. 
For some tasks, our model gives the best results with only 1/10 English training data.
 All codes are publicly available at: \url{https://github.com/kodenii/CoSDA-ML}.
\section{Background}
In this section, we will describe the background of mBERT as well as how to apply mBERT for cross-lingual classification tasks and sequence labeling tasks.

\subsection{mBERT}\label{mBERT}
mBERT follows the same model architecture and training procedure as BERT \cite{devlin-etal-2019-bert}.
It adopts a 12 layer Transformer,
but instead of training only on monolingual English data, it is trained on the Wikipedia pages of 104 languages with a shared word piece vocabulary, which allows the model to share embeddings across languages.
\subsection{Fine-tuning mBERT for Classification}
Given an input utterance $\textbf{s} = ({s_{1}, s_{2}, ..., s_{n}})$ from a source language (i.e., English), we first construct the input sequence by adding specific tokens  $\textbf{s} = ({\texttt{[CLS]}, s_{1}, s_{2}, ..., s_{n}, \texttt{[SEP]}})$, where $\texttt{[CLS]}$ is the special symbol for representing the whole sequence, and $\texttt{[SEP]}$ is the special symbol to separate non-consecutive token sequences \cite{devlin-etal-2019-bert}.
mBERT takes the constructed input sequence of no more than 512 tokens 
and outputs the representation of the sequence ${h}$ =  ($\boldsymbol{h}_{\texttt{CLS}}$, $\boldsymbol{h}_{1}$, \dots , $\boldsymbol{h}_{n}$, $\boldsymbol{h}_{\texttt{SEP}}$).

For classification tasks, mBERT takes $\boldsymbol{h}_\texttt{CLS}$ into a classification layer to find the label $c$:
\begin{align}
%p(c|{h})
c =\operatorname{softmax}(\boldsymbol{W}{\boldsymbol{h}}_{\texttt{CLS}} + \boldsymbol{b}),
\end{align}
where $\boldsymbol{W}$ is a task-specific parameter matrix. We fine-tune all the parameters of mBERT as well as $\boldsymbol{W}$ jointly by maximizing the log-probability of the correct label.

\subsection{Fine-tuning mBERT for Sequence Labeling}
For sequence labeling tasks, we feed the final hidden states of the input tokens into a softmax layer to classify the tokens. 
Note that BERT produces embeddings in the  wordpiece-level with WordPiece tokenization. We use the hidden state corresponding to the first sub-token as input to classify a word.
\begin{align}
\boldsymbol{y}_n = \operatorname{softmax}({ \boldsymbol{W}}^s \boldsymbol{h}_n + \boldsymbol{b}^s) \,, 
%n \in {1 \dots N}\
\end{align}
\noindent  where \(\boldsymbol{h}_n\) is the first sub-token representation of word \(x_n\).

\subsection{Zero-Shot Cross-Lingual Adaption}
 The baseline mBERT models, trained on the source language classification and sequence labeling tasks, perform \textit{zero-shot} cross-lingual transfer tasks by directly being used for the target language.
 We assume that there are labeled training data for each task in English, and transfer the trained model to a target language without labeled training data.

\begin{algorithm}[t]
	\KwIn{source language training data: 
		$\textbf{S}$ = $\{{\textbf{s}^{(n)}}\}_{n=1}^N$;
		a set of bilingual dictionaries: $dict$; 
		\{sentence, token\} replacement ratio: \{$\alpha$, $\beta$\};
		target language sets: $LAN$
	}
	\KwOut{multi-lingual code-switching training data: $\textbf{T} = \{\textbf{t}^{(n)}\}_{n=1}^N$.}
		\For{$n \gets 1 ... N$}{
		{
			\eIf{random() $ <\alpha$}{
				$i \gets 0$\;
					$ \textbf{t}^{(n)} \gets \emptyset$\; 
					\While{$s_i^{n} \not= \texttt{[SEP]}$}{
				\eIf{random() $ <\beta$}{
					$tgt \gets random(LAN)$\;
					$s_i^{n} \gets dict_{src}^{tgt}[s_i^{n}]$\;
				}{
					$s_i^{n} \gets s_i^{n} $\;
				}
				$\textbf{t}^{(n)} \gets \textbf{t}^{(n)}  \cup \{s_i^{n}\}$\;
				$i \gets i + 1$\;
			}
			}{
				$\textbf{t}^{(n)} \gets\textbf{s}^{(n)}$\;
			}
		}
	}
	\caption{Multi-lingual code-switching data augmentation framework.
	}\label{algo:generic}
\end{algorithm}
\section{Method}
Our method enhances mBERT in \S\ref{mBERT}.
 In this section, we first describe the overall training process (\S\ref{train_adaption}).
Then, we explain our augmentation algorithm in detail (\S\ref{augmentation_algorithm}).
Finally, we introduce the conducted tasks and their input construction for mBERT 
(\S\ref{tasks}).
\subsection{Training and Adaptation}\label{train_adaption}
\begin{figure}[t!]
	\centering
	\includegraphics[width=0.9\columnwidth]{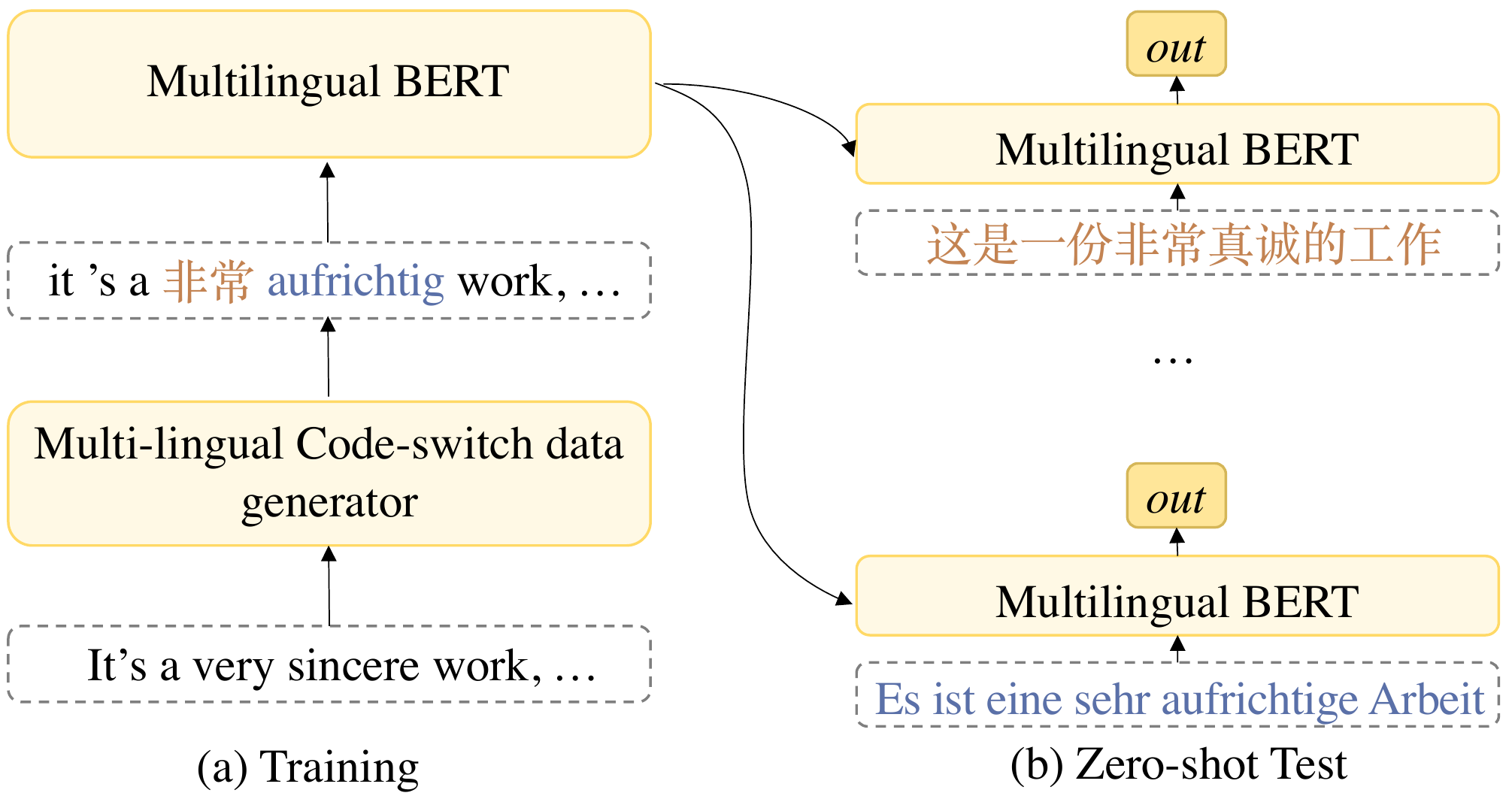}
	\caption{Illustration of our training and zero-shot test process with multi languages. Red color denotes Chinese and blue color denotes German. Better viewed in color.}\label{fig:framework}
\end{figure}

Our framework performs the cross-lingual tasks in two steps: Fine-tuning mBERT with augmented multi-lingual code-switch data and applying it for zero-shot testing, which is illustrated in Figure~\ref{fig:framework}.
Given a batch of training data $\textbf{S}$ = $\{{\textbf{s}^{(n)}}\}_{n=1}^N$ from a source language, the dynamic augmentation generator adopts Algorithm~\ref{algo:generic} to generate code-switched training data for fine-tuning mBERT.
Formally, the procedure can be written as:
\begin{eqnarray}
\mathbf{T} &=& \textnormal{Generator}(\mathbf{S}), \\
out &=& \textnormal{Fine-tune}(\textnormal{mBERT}, \mathbf{T}),
\end{eqnarray}
where $\mathbf{T}$ represents the generated code-switched data, $out$ denotes the output of all tasks.
In zero-shot test, the fine-tuned mBERT is used directly for target languages.
\subsection{Data Augmentation Algorithm}\label{augmentation_algorithm}
The augmentation method consists of three steps, including sentence selection, word selection and replacement selection.
\begin{itemize}
	\item[(i)] \textbf{Sentence Selection:} Given a batch of training data $\textbf{S}$, we randomly select sentences for  generating code-switched sentences.
	The unselected sentences keep in the original language. 
	Take the sentences in Figure~\ref{fig:example}(b) for example, we randomly select the first and the third sentence while leaving the second sentence unchanged;
	\item[(ii)] \textbf{Token Selection:} For each selected sentences in the sentence selection step, we randomly choose words to translate.
	Take the example in Figure~\ref{fig:example}(c). The word ``\textit{very}'' in first sentence and ``\textit{What}'' in third sentence are chosen;
	\item[(iii)] \textbf{Replacement Selection:} After obtaining the selected word, we randomly choose a target language according to a bilingual-dictionary.
	As shown in Figure~\ref{fig:example} (d), different target languages can be mixed in the code-switch generated data. It is worth noticing that words in the source language can have multiple translations in the target language. 
	In this case, we randomly choose any of the multiple translations as the replacement target language word.
	Though we cannot guarantee that this is the correct word-to-word translation in the context, we can consider it as one of the data augmented strategy
	 for our tasks.
\end{itemize}

Algorithm~\ref{algo:generic} shows pseudocode for the multi-lingual code-switching code augmentation process,
where lines 1-2 denote the sentence selection step,
lines 3-6 denote the word selection and lines 7-11 denote the replacement selection step.

In addition, the augmentation steps are performed per batch dynamically and the model trains with different augmented data in each batch, which can increase the distribution of data instances \cite{liu2019roberta}.
Intuitively, training with augmented code-switched data can make model automatically align the replaced word in the target language and the original word in a source language into a similar vector space according to their similar context information.
\subsection{Tasks}\label{tasks}
\paragraph{Natural Language Inference.}
We use XNLI \cite{conneau-etal-2018-xnli}, which covers 15 languages for natural language inference.
We feed a pair of sentences directly into the mBERT encoder and a task-specific classification layer is used for classification.
Models are evaluated by the classification accuracy (ACC).
\paragraph{Sentiment Classification.}
Following \newcite{barnes-etal-2018-bilingual}, we use the OpeNER English and Spanish datasets, and the MultiBooked Catalan and Basque datasets.
We directly provide the sentence to mBERT encoder and the specific \texttt{[CLS]} representation is fed into a linear layer for classification. 
 Models are evaluated by the macro F1.

\paragraph{Document Classification.}
We use MLDoc \cite{schwenk-li-2018-corpus} for document classification, which includes a balanced subset size of the Reuters corpus covering 8 languages for document classification.
Similar to sentiment classification, we also directly provide the document to mBERT encoder, and the specific \texttt{[CLS]} representation is fed into a linear layer for classification. 
Models are evaluated by classification accuracy (ACC).
\begin{table*}[!htbp]
	\begin{center}
		\renewcommand\arraystretch{1.1}
		\resizebox{\linewidth}{!}{
			\begin{tabular}{l|ccccccccccccccc|c}
				\hline
				\multirow{1}*{\textbf{Model}}
				& \textbf{en} & \textbf{fr} & \textbf{es} & \textbf{de} & \textbf{el} & \textbf{bg} & \textbf{ru} & \textbf{tr} & \textbf{ar} & \textbf{vi} & \textbf{th} & \textbf{zh} & \textbf{hi} & \textbf{sw} & \textbf{ur} & \textbf{Average} \\
				\hline
				\newcite{artetxe2018massively} & 73.9 & 71.9 & 72.9 & 72.6 & 73.1 & 74.2 & 71.5 & 69.7 & 71.4 & 72.0 & 69.2 & 71.4 & 65.5 & 62.2 & 61.0 & 70.2 \\
				XLM \cite{conneau2019cross} & 84.1 & 77.1 & 78.0 & 75.0 & 74.1 & 75.1 & 72.4 & 70.0 & 70.6 & 71.5 & 68.3 & 73.2 & 66.7 & 67.5 & 62.2 & 72.4 \\
				\quad+CoSDA-ML  & \textbf{84.4*} & \textbf{79.0*} & \textbf{79.2*} & \textbf{77.9*} & \textbf{76.8*} & \textbf{77.6*} & \textbf{75.7*} & \textbf{72.6*} & \textbf{73.4*} & \textbf{75.3*} & \textbf{72.6*} & \textbf{75.1*} & \textbf{71.2*} & \textbf{70.0*} & \textbf{68.3*} & \textbf{75.3*} \\
				\hdashline
				mBERT from \newcite{wu-dredze-2019-beto} & 82.1 & 73.8 & 74.3 & 71.1 & 66.4 & 68.9 & 69.0 & 61.6 & 64.9 & 69.5 & 55.8 & 69.3 & 60.0 & 50.4 & 58.0 & 66.3 \\
				\quad+CoSDA-ML  & {82.9} & {76.7} & {76.9} & {74.1} & {70.9} & {72.7} & {73.2} & {63.9} & {68.0} & {73.6} & {59.8} & {73.8} & {65.5}& {51.0} & {62.3} & {69.7} \\
				\hline
			\end{tabular}
		}
		\caption{Natural Language Inference experiments. 
		}
		\label{tab:xnli}
	\end{center}
\end{table*}

\begin{table*}[!htbp]
	\small
	\centering
	\renewcommand\arraystretch{1.1}
	\resizebox{0.65\linewidth}{!}{
		\begin{tabular}{l|c|c|c|c|c|c} 
			\hline 
			\multirow{2}*{\textbf{Model}} & \multicolumn{2}{c|}{\textbf{es}} & \multicolumn{2}{c|}{\textbf{eu}} & \multicolumn{2}{c}{\textbf{ca}} \\ 
			\cline{2-7} 
			~ & \multicolumn{1}{c|}{2-Class} & \multicolumn{1}{c|}{4-Class} & \multicolumn{1}{c}{2-Class} & \multicolumn{1}{|c}{4-Class} & \multicolumn{1}{|c}{2-Class} & \multicolumn{1}{|c}{4-Class} \\ 
			\cline{2-5}  
			\hline   
			BLSE \cite{barnes-etal-2018-bilingual}  & 74.6 & 41.2 & 69.3 & 30.0 & 72.9 & 35.9 \\ 
			\hline
			XLM \cite{conneau2019cross}  & 86.1 & 32.9 & - & - & - & -  \\ 
			\quad+CLCSA   & 91.3 & 46.8 & - & - & - & -  \\  
			\hdashline
			mBERT \cite{devlin-etal-2019-bert}  & 93.1 & 51.0 & 73.1 & 35.1 & 83.5 & 52.3  \\ 
			\quad+CLCSA   & \textbf{95.2*} & \textbf{57.9*} & \textbf{85.7*} & \textbf{55.6*} &\textbf{ 88.7*} & \textbf{64.3*}  \\

			\hline  
		\end{tabular}
	}
	\caption{Sentiment classification experiments. } 
	\label{table:sc}
	
\end{table*}

\paragraph{Dialogue State Tracking (DST).}
Following prior work \cite{liu2019attentioninformed}, we use the Multilingual WOZ 2.0 dataset \cite{mrksic-etal-2017-semantic}, which includes German and Italian languages.
DST aims to predict the slot-value pair given a current utterance and the previous system acts.
It can be viewed as a collection of binary prediction problems by using a distinct estimator for each slot-value pair \cite{chen-etal-2018-xl}.
We concatenate the current utterance and the previous system act for input into mBERT and obtain the \texttt{[CLS]} representation.
We also feed each slot-value pair into mBERT and obtain another \texttt{[CLS]} representation.
Finally, the two representations are provided to the classification layer to decide whether it should be selected.
Similar to prior work, we use the turn-level request tracking accuracy, joint goal tracking accuracy and the slot tracking accuracy for evaluation.
\paragraph{Spoken Language Understanding.}
We follow \newcite{schuster-etal-2019-cross} and use the cross-lingual spoken language understanding dataset, which contains English, Spanish and Thai. 
We adopt a joint model which  provides the utterance for mBERT and the \texttt{[CLS]} is used for intent detection.
The token representations are used for slot prediction as local classifier task on each word, which can be treated as a sequence labeling task.
Intent detection is evaluated by the classification accuracy (ACC) and slot filling is evaluated by F1 score.
%\cite{zhong-etal-2018-global}

\section{Experiments}
We evaluate the effectiveness of our proposed method across 19 languages on five tasks.
In addition to mBERT, we also conduct all experiments on the recent strong pre-trained cross-lingual model (XLM) \cite{conneau2019cross}.
XLM outperforms mBERT on XNLI tasks, but underperforms mBERT for some other tasks \cite{liu2019attentioninformed}.
We choose it as a secondary baseline for verifying the generalizability of our augmentation method.
\subsection{Experimental Settings}
For all tasks, no preprocessing is performed except tokenization of words into subwords with WordPiece.
Following \newcite{devlin-etal-2019-bert}, we use WordPiece embeddings with a 110k token vocabulary.
We use the base case multilingual BERT (mBERT), which has N = 12 attention heads and M = 12 Transformer blocks. 
In fine-tuning, we select the best hyperparameters by searching a combination of batch size, learning rate, the number of fine-tuning epochs and replacement ratio with the following range: learning rate $\{1\times 10^{-6},2\times 10^{-6},3\times 10^{-6},  4\times 10^{-6},  5\times 10^{-6},1\times 10^{-5}\}$; batch size $\{8, 16, 32\}$; number of epochs: $\{4, 10, 20, 40, 100\}$;
token and sentence replacement ratio: $\{0.4, 0.5, 0.6, 0.8, 0.9, 1.0\}$. 
Note that the best model are saved by development performance in the \textit{English}.

\subsection{Baselines}
We include the following state-of-the-art baselines:

\paragraph{Natural Language Inference.}\label{nli}
\newcite{artetxe2018massively} use multilingual sentence representation, pre-trained with sequence-to-sequence model.
 This model requires bitext for training. 

\paragraph{Sentiment Classification.}
BLSE \cite{barnes-etal-2018-bilingual} jointly represents sentiment information in a source and target language and achieves the state-of-the-art performance in zero-shot cross-lingual sentiment classification.
\begin{table*}[!htbp]\label{MLDoc}
 \begin{center}
  \renewcommand\arraystretch{1.2}
  \resizebox{0.8\linewidth}{!}{
   \begin{tabular}[b]{l|ccccc ccc|c}
    \hline
    \textbf{{Model}} & \textbf{en} & \textbf{de} & \textbf{zh} & \textbf{es} & \textbf{fr} & \textbf{it} & \textbf{ja} & \textbf{ru} & \textbf{Average} \\
    \hline
    
    \hline
    \newcite{schwenk-li-2018-corpus} & {92.2}& {81.2} &{74.7} & 72.5 & 72.4 & {69.4} & {67.6} & 60.8 & 73.9 \\
    \newcite{artetxe2018massively}  & {89.9}& {84.8} & 71.9 & {77.3} & {78.0} & {69.4} & {60.3} & {67.8} & {74.9} \\
    \hline
    XLM \cite{conneau2019cross}   & 94.2 & 76.8 & 46.2 & 64.0 & 70.5 & - & - & 61.5 & 68.9 \\
    \quad+CLCSA   & 93.4 & 81.4& 71.1 & 73.1 & 83.7 & - & - & 68.3 & 78.5 \\
    \hdashline
    mBERT \cite{devlin-etal-2019-bert}   & {94.2}& 80.2 & 76.9 & 72.6 & 72.6 & 68.9 & 56.5 & 73.7 & 74.5 \\
    \quad+CLCSA   & \textbf{95.2*} & \textbf{86.3*}& \textbf{{85.5*}} & \textbf{{79.2*}} & \textbf{86.7*} & \textbf{{72.6*} }& \textbf{73.7*} & \textbf{{75.1*}} & \textbf{{81.8*}} \\
    \hline
   \end{tabular}
  }
  \caption{Document classification experiments. 
  }
  \label{tab:mldoc}
 \end{center}
 \vspace{-0.4cm}
\end{table*}

\begin{table*}[!htbp]
	\small
	\centering
	\renewcommand\arraystretch{1.1}
	\begin{adjustbox}{width=0.8\textwidth}
		\begin{tabular}{l|c|c|c|c|c|c} 
			\hline 
			\multirow{2}*{\textbf{Model}} & \multicolumn{3}{c|}{\textbf{German}} & \multicolumn{3}{c}{\textbf{Italian}} \\ 
			\cline{2-7} 
			~ & \multicolumn{1}{c|}{slot acc.} & \multicolumn{1}{c|}{joint goal acc.} & \multicolumn{1}{c}{request acc.} & \multicolumn{1}{|c|}{slot acc.} & \multicolumn{1}{c|}{joint goal acc.}& \multicolumn{1}{c}{request acc.} \\ 
			\cline{2-7}  
			\hline  
			XL-NBT  \cite{chen-etal-2018-xl}   & 55.0 & 30.8 & 68.4& 72.0 & 41.2 & 81.2\\ 
			Attention-Informed Mixed Training \cite{liu2019attentioninformed}   & 69.5 & 32.2 & 86.3 & 69.5 & 31.4 & 85.2\\ 
			\hline 
			XLM from \newcite{liu2019attentioninformed}  & 58.0 & 16.3 & 75.7 & - & - & -  \\ 
			\quad+CLCSA  & {77.4} & {48.7} & {88.3} & - & - & -  \\  
			\hdashline
			mBERT \cite{devlin-etal-2019-bert}  & 57.6 & 15.0 & 75.3 & 54.6 & 12.6 & 77.3  \\ 
			\quad+CLCSA  & \textbf{83.0*} & \textbf{63.2*} & \textbf{94.0*} & \textbf{82.2*} & \textbf{61.3*} & \textbf{94.2*}  \\

			\hline  
		\end{tabular}
	\end{adjustbox}
	\caption{Dialog State Tracking experiments. } 
	\label{table:dst}
	
\end{table*}

\begin{table*}[!htbp]
	\small
	\centering
	\renewcommand\arraystretch{1.1}
	\resizebox{0.6\linewidth}{!}{
		\begin{tabular}{l|c|c|c|c} 
			\hline 
			\multirow{2}*{\textbf{Model}} & \multicolumn{2}{c|}{\textbf{Spanish}} & \multicolumn{2}{c}{\textbf{Thai}} \\ 
			\cline{2-5} 
			~ & \multicolumn{1}{c|}{Intent acc.} & \multicolumn{1}{c|}{Slot F1} & \multicolumn{1}{c}{Intent acc.} & \multicolumn{1}{|c}{Slot F1} \\ 
			\cline{2-5}  
			\hline 
			Multi. CoVe \cite{yu-etal-2018-multilingual}  & 53.9 & 19.3 & 70.7 & 35.6 \\ 
			Attention-Informed Mixed Training \cite{liu2019attentioninformed}   & 86.5 & 74.4 & 70.6 & 28.5 \\ 
			\hline 
				XLM from \newcite{liu2019attentioninformed}  & 62.3 & 42.3 & 31.6 & 7.9 \\ 
				\quad+ CLCSA   & {90.3} &{69.0} & \textbf{86.7} & {34.9} \\  
				\hdashline
				mBERT \cite{devlin-etal-2019-bert}  & 73.7 & 51.7 & 28.2& 10.6 \\ 
				\quad+ CLCSA  (Static)   & 92.8 & 75.2 & 74.8 & 28.1 \\
				\quad+ CLCSA   & \textbf{94.8*} & \textbf{80.4*} & {76.8} & \textbf{37.3*} \\

			\hline  
		\end{tabular}
	}
	\caption{Slot filling and Intent detection experiments.} 
	\label{table:slu}
	
\end{table*}

\paragraph{Document Classification.}
1) \newcite{schwenk-li-2018-corpus} use MultiCCA, multilingual word embeddings trained with a bilingual dictionary, and convolution neural networks. 2) \newcite{artetxe2018massively} also obtain the promising performance and the detail has been described in Natural Language Inference paragraph.
%~\ref{nli}
\paragraph{Dialogue State Tracking (DST).}
%We include two strong baselines:
1) XL-NBT \cite{chen-etal-2018-xl} proposes a state tracker for the source language as a teacher and then distills and transfers its own knowledge to the student state tracker in target languages. 
2) {Attention-Informed Mixed Training}: \newcite{liu2019attentioninformed} use the generated attention-informed code-switch data for training and achieves the state-of-the-art performance.
\paragraph{Spoken Language Understanding.}
1) {Multi. CoVe}: \cite{schuster-etal-2019-cross-lingual} use Multilingual CoVe \cite{yu-etal-2018-multilingual} as the encoder and add an  autoencoder objective to produce more general representations for semantically similar sentences across languages.
2) {Attention-Informed Mixed Training}: \newcite{liu2019attentioninformed} use attention to generate code-switched sentences, achieving the current best result. The method translates only one word into each augmented sentence.
\subsection{Results}
We perform t-test for all experiments to measure whether the results from the proposed model are significantly better than the baselines. The numbers with asterisks indicate that the improvement is significant with $p < 0.01$. ``-'' represents the absence of languages in the XLM models and we cannot report the results.
Five tasks results are shown in Table~\ref{tab:xnli}, ~\ref{table:sc}, ~\ref{tab:mldoc}, ~\ref{table:slu} and ~\ref{table:dst}, respectively.
Across the tasks, we can observe that:
1) mBERT achieves strong performance on all zeros-shot cross-lingual tasks, which demonstrates that mBERT is a surprisingly effective cross-lingual model for a wide range of NLP tasks.
This is consistent with the observation of \newcite{wu-dredze-2019-beto}.  Additionally, the XLM achieves much better performance than mBERT on XNLI and achieves the promising performance on four other tasks.
2) Our method outperforms mBERT and XLM by a large margin and achieves state-of-the-art performance on all the tasks, which demonstrates the effectiveness of our proposed method.
Note that we have not reproduced the results on XNLI task of original paper because of lacking the exact best hyper-parameters, which is also mentioned on some issues on Github.\footnote{https://github.com/facebookresearch/XLM/issues/199.} 
So we run their open-source code \footnote{https://github.com/facebookresearch/XLM} to obtain the results and we apply the CoSDA-ML to it with the same hyper-parameters.
Besides, our method not only obtains 2.9\% improvement on average score but also outperforms the reported results (Average 75.1 score) from \newcite{conneau2019cross}, which further demonstrates the effectiveness of our method.
3) Our method outperforms \textit{Attention-Informed Mixed Training} in both DST and SLU tasks, which indicates that our dynamic sampling and multi-lingual code-switch data training technique are more effective for aligning representation between source and target languages than only translating one word to the target language.
\begin{figure*}[!ht]
	\centering
	\begin{subfigure}{.35\textwidth}
		\centering
		\includegraphics[scale=0.55]{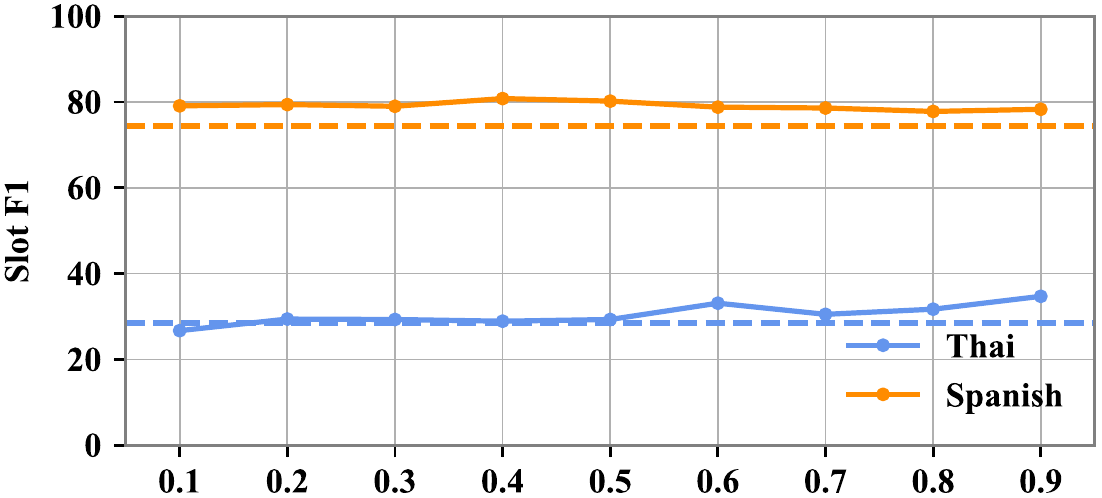}
		\caption{Slot filling.}
		\label{fig:intent-es}
	\end{subfigure}
	\begin{subfigure}{.35\textwidth}
		\centering
		\includegraphics[scale=0.55]{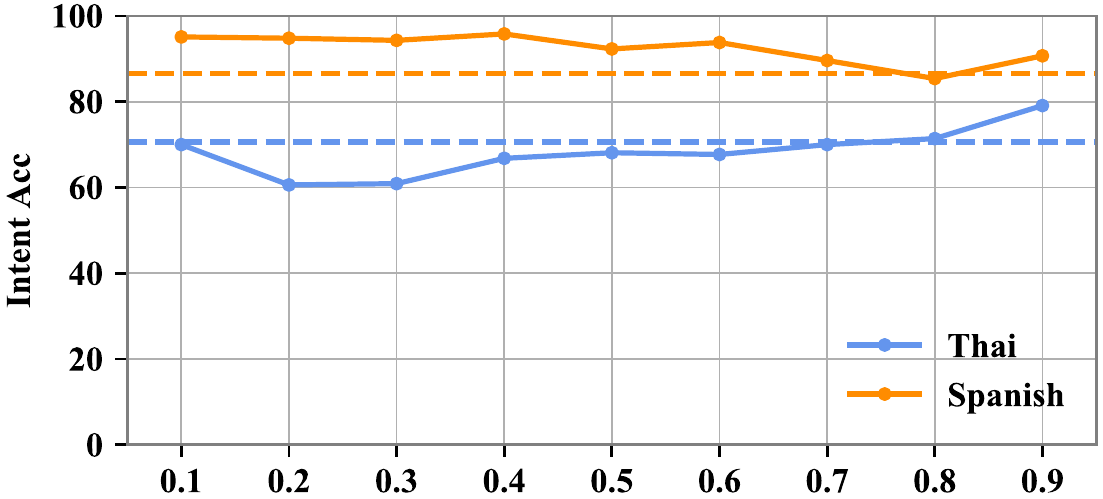}
		\caption{Intent detection.}
		\label{fig:slot-es}
	\end{subfigure}
	\begin{subfigure}{.35\textwidth}
		\centering
		\includegraphics[scale=0.55]{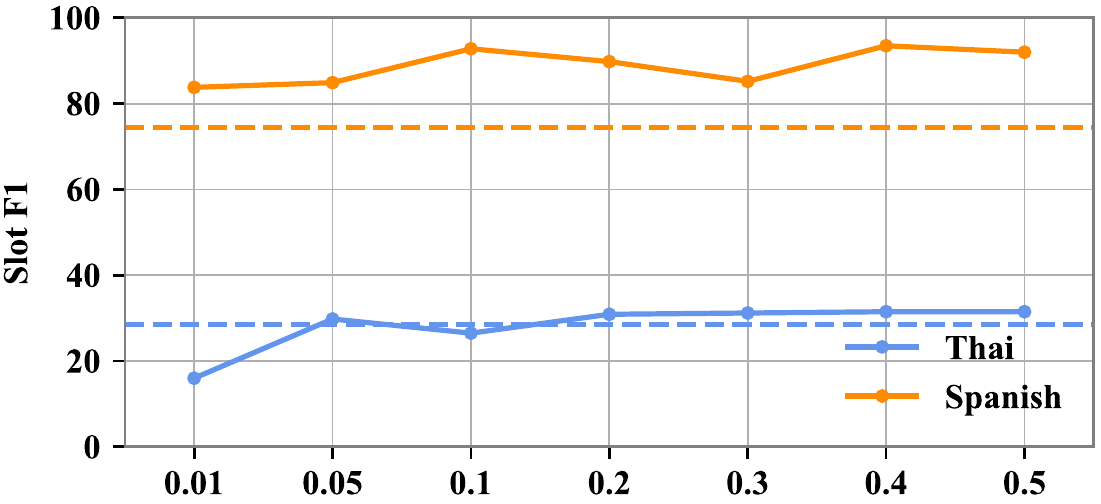}
		\caption{Slot filling.}
		\label{fig:intent-es-unseen}
	\end{subfigure}
	\begin{subfigure}{.35\textwidth}
		\centering
		\includegraphics[scale=0.55]{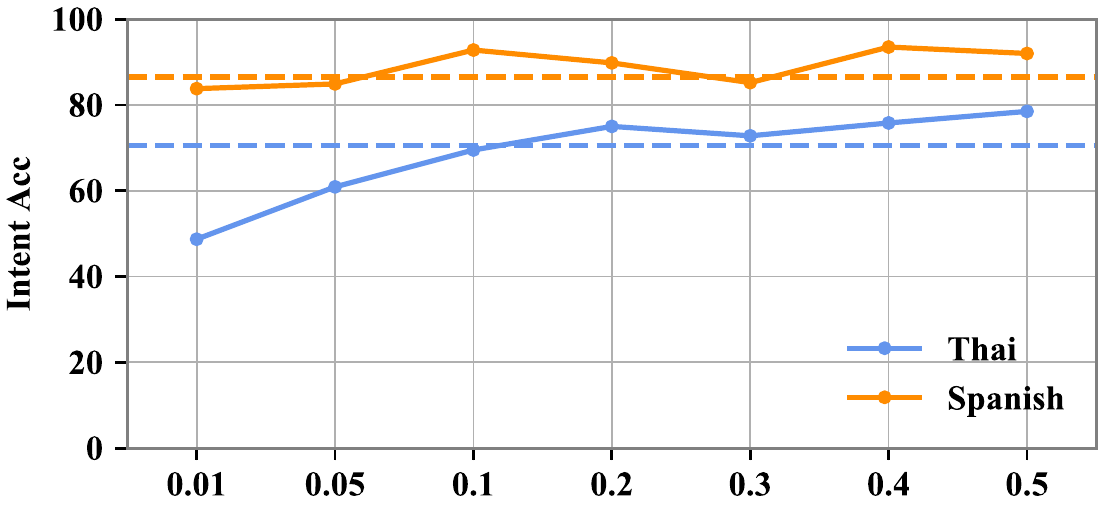}
		\caption{Intent detection.}
		\label{fig:slot-es-unseen}
	\end{subfigure}
	\caption{Comparison between our model (solid lines) and Attention-Informed Mixed Training (Att.) model (dashed lines). Results with different $\beta$ in (a) and (b) and different subset size of training data in (c) and (d). 
		In (c) and (d), it's worth that the dashline denotes Att. performance with 100\% training data and the solid line represents our model performance by varying the proportion training data size.}
	\label{fig:dynamics}
\end{figure*}
\begin{figure}[t!]
	\centering
	\begin{subfigure}[t]{0.2\textwidth}
		\centering
		\includegraphics[height=0.8in]{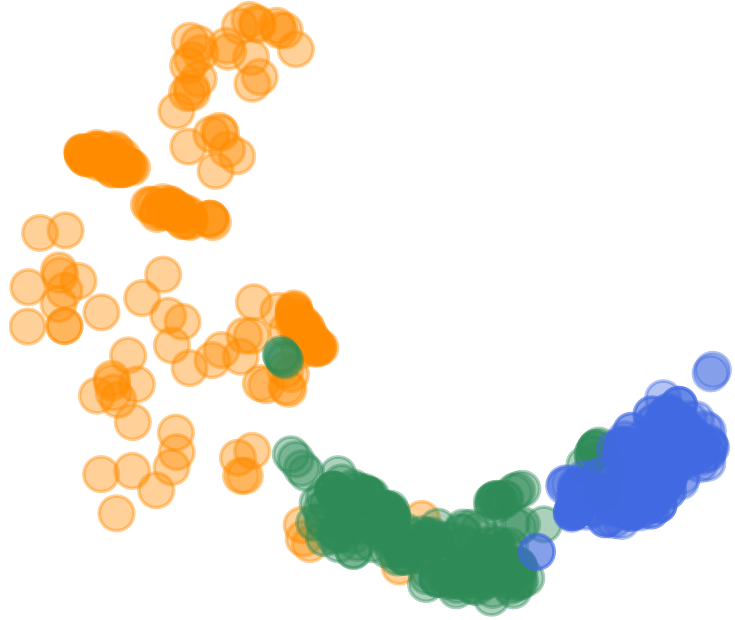}
		\caption{}
	\end{subfigure}%
	~ 
	\begin{subfigure}[t]{0.2\textwidth}
		\centering
		\includegraphics[height=0.8in]{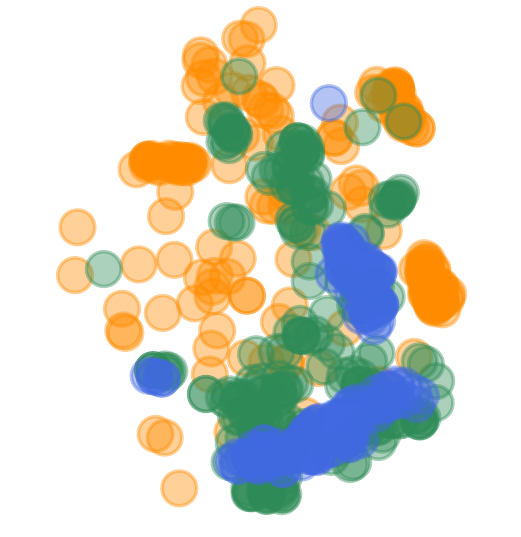}
		\caption{}
	\end{subfigure}
	\caption{t-SNE visualization of sentences vector space from mBERT (a) and with our CoSDA-ML method (b). The different color represents different languages and the dots in the same color denotes sentence representation with same intent.}
	\label{fig:vis}
\end{figure}

\subsection{Analysis}

\paragraph{Robustness.}
To verify the robustness of CoSDA-ML, we conduct experiments with different token replacement ratios $\beta$ during the fine-tuning process and keep the sentence replacement ratio $\alpha$ as 1.
The results are shown in Figure \ref{fig:dynamics}(a) and \ref{fig:dynamics}(b).
With all the values of $\beta$,  our model consistently outperforms the state-of-the-art model (\textit{Attention-Informed Mixed Training}) in slot filling and intent detection when $\beta$ $>$ 0.7,  which verifies the robustness of our method.

\paragraph{Varying Amounts of Training Data.}
We study the effectiveness of CoSDA-ML by varying amounts of training data.
Figure~\ref{fig:dynamics}(c) and ~\ref{fig:dynamics}(d) report the results
of adding varying amounts of training data between \textit{Attention-Informed Mixed Training} and our model.
We have two interesting observations:
1)  Our augmentation method consistently outperforms the baseline with all training data sizes, which demonstrates consistency.
2) Using only 1/10 of the training data,
our approach performs better than the \textit{Attention-Informed Mixed Training} using 100\% of the training data, demonstrating that
our approach is particularly useful when we only access to small amounts of training data.

\paragraph{Effectiveness of Dynamic Sampling.}
To verify the effectiveness of our proposed dynamic augmentation mechanism, we make comparison with \textit{static} augmentation method, in which we adopt Algorithm~\ref{algo:generic} to obtain augmented multi-lingual code-switch training data once for all the batches.
The results are shown in the \textit{static} row of Table~\ref{table:slu} .
We find that the dynamic method outperforms the \textit{static} method in all the tasks.
We attribute this to the fact that the dynamic mechanism can generate more varying code-switched multi-lingual data within the batch training process while \textit{static} method can only augment one time of origin training data. Dynamic sampling allows the model to align more words representation closer in multiple languages.

\begin{figure}
	\centering
	\includegraphics[width=0.85\linewidth]{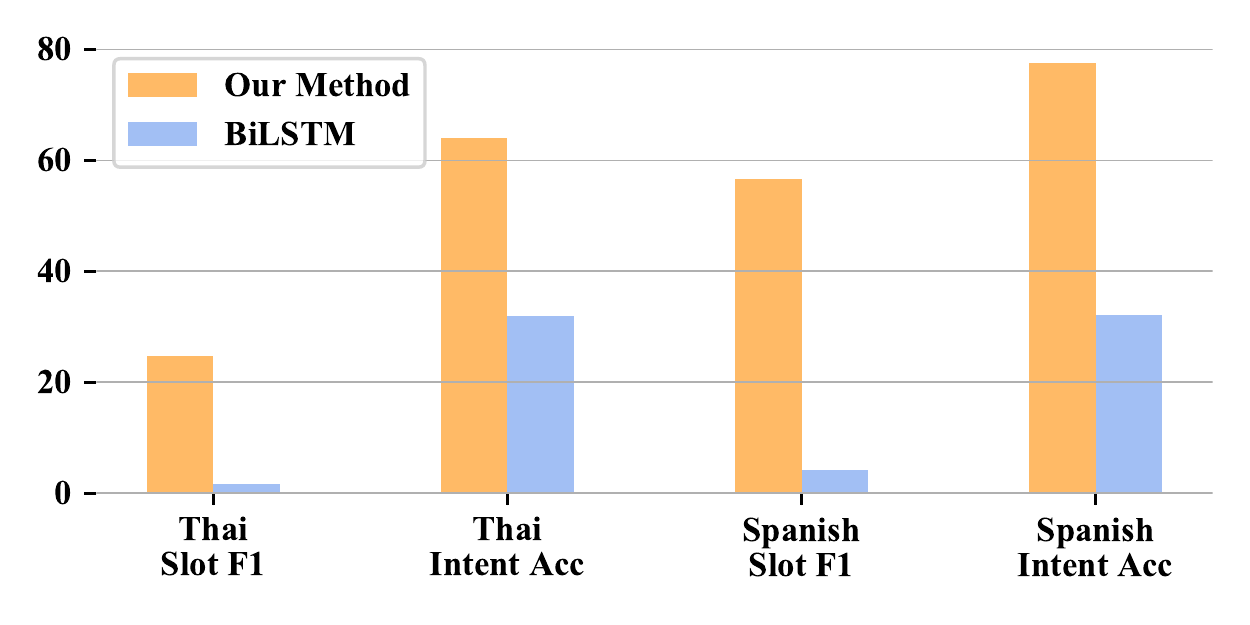}
	\caption{Evaluation result based on BiLSTM.}
	\label{fig:bilstm}
\end{figure}

\paragraph{Visualization.}
In order to see whether our framework aligns the representation between the source language and all the target languages,
we select three intents with 100 sentences respectively and obtain their sentence vector $\texttt{[CLS]}$ to visualize between our method with mBERT. 
The mBERT results are shown in Figure~\ref{fig:vis}(a).
We can see that there is nearly no overlap between different languages, which shows that the distance of the representations of different languages with the same intent is distant.
In contrast, the representations from our CoSDA-ML fine-tuned model in Figure~\ref{fig:vis} (b) in different languages become closer and overlap with each other, which further demonstrates that our method effectively and successfully aligns representations of different languages closer.
\paragraph{CoSDA-ML with BiLSTM.}
A natural question that arises is whether our augmentation method is effective for a general encoder in addition to Transformer.
To investigate the question,
we replace mBERT with BiLSTM and keep other components the same.
BiLSTM does not include any information pre-trained over  Wikipedia pages of multiple languages.
We conduct experiments on top of BiLSTM to better verify whether our method strongly depends on the pre-trained model.
The results are shown in Figure~\ref{fig:bilstm}.
We can see that our framework outperforms BiLSTM in all metrics in all languages, which further demonstrates that our augmentation method is not only effective on top of mBERT but also can work on a general encoder.
\section{Related Work}
\paragraph{Zero-shot Cross-lingual Transfer.}
The main strands of work focused on learning \textit{cross-lingual word embeddings}. 
\newcite{ruder2017survey} surveyed methods
 \cite{klementiev2012inducing,kocisky-etal-2014-learning,guo2016representation}
  for learning cross-lingual word embeddings by either joint training or post-training mappings of monolingual embeddings.
\newcite{xing2015normalized}, \newcite{lample2018word} and
 \newcite{chen2018unsupervised} proposed to take pre-trained monolingual word embeddings of different languages as input, aligning them into a shared semantic space.
Our work follows in the recent line of cross-lingual contextualized embedding methods \cite{huang-etal-2019-unicoder,devlin-etal-2019-bert,wu-dredze-2019-beto,conneau2019cross,artetxe2019cross}, which are trained using masked language modeling or other auxiliary pre-training tasks to encourage representation in source and target language space closer, achieving state-of-the-art performance on a variety of 
 zero-shot cross-lingual NLP tasks.
 In addition, our work is related with the recent work \cite{conneau-etal-2020-emerging},which analyzed the effectiveness of anchor points. 
We propose a data augmentation framework to dynamically construct multi-lingual code-switching data for training, which encourages model implicitly to align similar words in different languages into the same space.

\paragraph{Data Augmentation.}
Recently, some augmentation methods have been successfully applied in the cross-lingual setting.
 \newcite{liu2019attentioninformed} proposed an attention mechanism to select the most important word to translate into the target language for training.
In contrast, our framework can augment data dynamically in each epoch to encourage the model to align the representation in different languages, and can generate multiple languages code-switch data making training once and directly testing for all languages multiple times.
\newcite{zhang-etal-2019-cross} proposed using code-mixing to perform the syntactic transfer in dependency parsing.
However, they need a high-accuracy translator to obtain multiple language data which can be difficult to train for low-resource language.
In contrast, our method uses the existing bilingual dictionaries, which can be more practical and useful.

\section{Conclusion}
We proposed an augmentation framework to generate multi-lingual code-switching data to fine-tune mBERT for aligning representations from source and multiple target languages.
Experiments on five tasks show that our method consistently and significantly outperforms mBERT and XLM baselines.
In addition, our method is flexible and can be used to fine-tune all base encoder models.
Future work includes the application of CoSDA-ML on the task of multi-lingual language modeling task, so that a more general version of the multi-lingual contextual embedding can be investigated.

\section*{Acknowledgements}
This work was supported by the National Natural Science Foundation of China (NSFC) via grant 61976072, 61632011 and 61772153. Besides, this work also faxed the support via Westlake-BrightDreams Robotics research grant.
We thank Yijia Liu for the helpful discussion and anonymous reviewers for the insightful comments. Wanxiang Che and Yue Zhang are the corresponding author. 
\bibliographystyle{named}
\bibliography{ijcai20}

\end{document}